\documentclass[conference]{IEEEtran}
\IEEEoverridecommandlockouts
\usepackage{cite}
\usepackage{amsmath,amssymb,amsfonts}
\usepackage{algorithmic}
\usepackage{graphicx}
\usepackage{textcomp}
\usepackage{xcolor}
\usepackage{booktabs}
\usepackage{enumitem}
\usepackage{floatrow}
\usepackage{bm}
\usepackage{subfigure}
\usepackage{caption}
\usepackage[square,sort,comma,numbers]{natbib}

\def\BibTeX{{\rm B\kern-.05em{\sc i\kern-.025em b}\kern-.08em
    T\kern-.1667em\lower.7ex\hbox{E}\kern-.125emX}}
\begin{document}

\title{Learning Syntactic and Dynamic Selective Encoding for Document Summarization
}

\author{\IEEEauthorblockN{Haiyang Xu, Yahao He, Kun Han, Junwen Chen and Xiangang Li}
\IEEEauthorblockA{\textit{AI Labs} - \textit{Didi Chuxing Co., Ltd.} - Beijing, China\\ 
Email: \{xuhaiyangsnow, heyahao, kunhan, chenjunwen, lixiangang\}@didiglobal.com}
}

\maketitle

\begin{abstract}
Text summarization aims to generate a headline or a short summary consisting of the major information of the source text. Recent studies employ the sequence-to-sequence framework to encode the input with a neural network and generate abstractive summary. However, most studies feed the encoder with the semantic word embedding but ignore the syntactic information of the text. Further, although previous studies proposed the selective gate to control the information flow from the encoder to the decoder, it is static during the decoding and cannot differentiate the information based on the decoder states. In this paper, we propose a novel neural architecture for document summarization. Our approach has the following contributions: first, we incorporate syntactic information such as constituency parsing trees into the encoding sequence to learn both the semantic and syntactic information from the document, resulting in more accurate summary; second, we propose a dynamic gate network to select the salient information based on the context of the decoder state, which is essential to document summarization. The proposed model has been evaluated on CNN/Daily Mail summarization datasets and the experimental results show that the proposed approach outperforms baseline approaches.
\end{abstract}


\section{Introduction}
\label{sec:intro}
Text summarization is a very challenging task of natural language processing (NLP) and information retrieval. Existing approaches for text summarization are categorized into two major types: extractive and abstractive. Extractive methods produce summaries by extracting sentences or tokens from the source text, which can produce the grammatically correct summaries and preserve the meaning of the original text. However, these approaches heavily rely on the text in the original documents and the extracted sentences may contain redundant information or be lack of readability. On the contrary, abstractive methods produce the summaries by generating new sentences or tokens which do not necessarily appear in the source text. 
%which aim to generate high-quality and more like human-written summaries. 
However, abstractive approaches are more difficult in practice because they need to address many NLP problems including document understanding, semantic representation and natural language generation, which are harder than sentence extraction.

\begin{figure}[h]
	%\vspace{-2ex}
	\centering
	\includegraphics[height=0.84\linewidth,width=\linewidth]{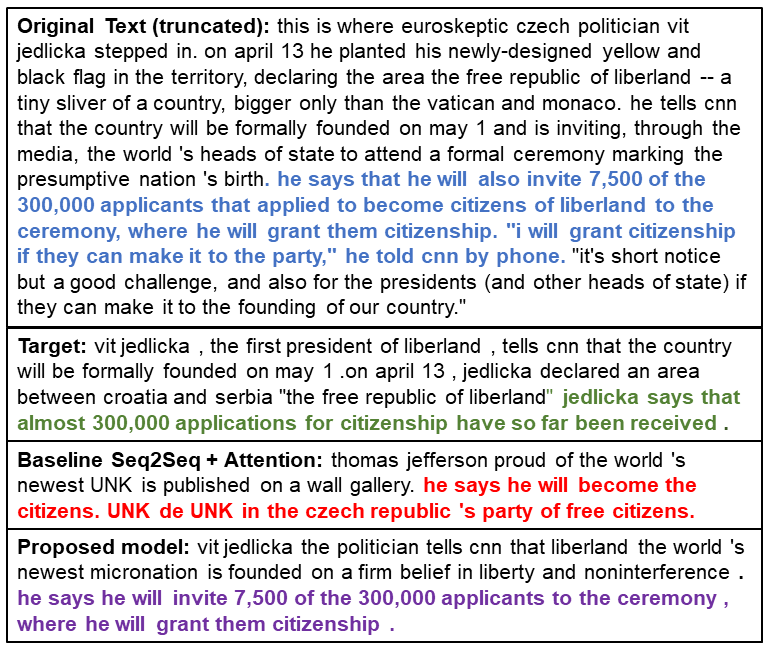}
	%\vspace{-2ex}
	\caption{The example of text summarization in CNN dataset. The colored text show the source text, corresponding summaries generated by human-written, original Seq2Seq with attention, and the proposed approach, respectively.}
	%\vspace{-2ex}
	\label{fig:example}
\end{figure}

The recent neural sequence-to-sequence (Seq2Seq) approach \cite{sutskever2014sequence} has achieved tremendous success in many NLP tasks such as machine translation \cite{bahdanau2014neural,luong2015effective}, dialogue systems \cite{serban2016building,bordes2016learning}. The essence of Seq2Seq based text summarization methods is an encoder-decoder framework, which first encodes a input sentence to a low dimensional representation and then decodes the abstract representation into a output sequence. As the extension of Seq2Seq methods, attention based Seq2Seq models encode a input sequence to a context vector using attention mechanism and dynamically calculate the attentive probability distribution at each generation step \cite{gu2016incorporating}. 

Similar to machine translation, some researchers have applied neural Seq2Seq model to abstractive text summarization \cite{nallapati2016abstractive,see2017get,tan2017abstractive}. However, there is a significant difference between two tasks: in machine translation, one aims to capture \textit{all} the semantic details from the source text, while in text summarization one only focuses on the \textit{salient} text information and it is critical to utilize only the key information in the source text rather than the whole text.

 %To address this problem, we utilize a dynamic selective gate to encode the salient information from the source text. 
Furthermore, the original Seq2Seq with attention method does not learn the syntactic structure of the source text, which is important to text summarization. In Figure~\ref{fig:example}, a piece of original text is shown at the top. The human-written summary is shown in the next box as the target. The third box shows a model-generated summary using the original Seq2Seq with attention method (baseline).  The green text in the target and the red text in the baseline show the summary corresponding to the blue text in the original text. As shown in the figure, the baseline model incorrectly summarizes that ``\textit{he says he will become the citizens}", mainly because it is not able to capture the internal syntactic structure of the original text (e.g. ``300, 000 applicants" is a noun phase, and ``applied to .. ceremony" is an attributive clause for it). 
 
%  Some researchers have applied neural Seq2Seq model to abstractive text summarization \cite{nallapati2016abstractive}\cite{see2017get}\cite{tan2017abstractive}.

%These two tasks have a big difference that the output sequence of translation needs to keep full semantic information of source text, while for text summarization the key idea is to select/generate salient information from source and filters out unnecessary sentences. 
%Furthermore, machine translation usually does not explicitly models the syntactic structure due to the different syntax across languages, but the syntactic information may be critical to text summarization. 

%, which has achieved huge success in machine translation \cite{eriguchi2016tree}\cite{li2017modeling}. 
%Syntactic information is important for phrase feature detection (Noun Phrase, Verb Phrase) and semantic understanding. 

%\begin{figure*}[h]
%	\includegraphics[height=3in,width=3.7in]{./figure/example}
%	\caption{The example of Baseline Seq2Seq + Attention, making factual errors, repeat, useless sentences and facing OOV problems.}
%	\label{fig:example}
%\end{figure*}

To address these problems, we propose a novel syntactic and dynamic selective encoding method for document summarization. We incorporate structured linguistic information such as parsing trees to learn more effective sentence representation by serializing parsing trees into a encoder sequence as in \cite{li2017modeling}. In this way, the encoding sequence contains both the semantic (words) and the syntactic (parsing symbols) information, both of which are fed into the decoder for summary generation.

In addition, a document may contains hundreds of words and it is hard to directly encode the key information from the whole source text. A selective gate was proposed in previous study \cite{Zhou2017Selective} to filter out the secondary information. However, the salient information varies in different decoding stage, so it is better to select the salient information based on the context of decoder states. Therefore, for each decoding step, we take a dynamic selective gate network to control the salient information flow according to the document information, current decoder state and the previous gated encoder state. 

%With the syntactic encoding and the dynamic selective gate, we incorporate the pointer-generator network \cite{vinyals2015pointer} and the coverage loss \cite{see2017get} for decoding to handle the out-of-vocabulary (OOV) problem and the word repetition.

%which can improve performance of word prediction and discourage word repetition.
% to learn useful syntactic information automatically; 
In this way, our approach can learn better representation of the sentences and select the salient information from the long input sequence for summary generation. As an example, we show the summary generated by our approach in Figure \ref{fig:example}, where the text correctly summarize the original sentences from the document. For reference, Figure \ref{fig:syntactic} shows the constituency parsing tree for the source sentence, and Figure \ref{fig:selective} shows the change of the states of the dynamic selective gates in a input sentence. We also conduct experiments on two large-scale CNN/Daily Mail datasets and the experimental results show that our model achieves superiority over baseline summarization models.

We organize the paper as follows: Sec.\ref{sec:related_work} introduces the related work. Sec.\ref{sec:method} describes our proposed method. In Sec.\ref{sec:exp} we present the experiment settings and illustrate experimental results. We conclude the paper in Sec.\ref{sec:con_fut}.

\section{Related Work}
\label{sec:related_work}
In general, there are two broad approaches to automatic text summarization: extractive and abstractive. Extractive methods work by selecting important sentences or passages in the original text and reproducing them as summary \cite{Wong2008Extractive,Wang2009Multi,celikyilmaz2010hybrid,Alguliev2011MCMR,Erkan2011LexRank,Alguliev2013Multiple}.

In contrast, abstractive summarization techniques generate new shorter texts that seldom consist of the identical sentences from the document. Banko et al. \cite{banko2000headline} apply statistical models of the term selection
and term ordering processes to produce short summaries. Bonnie and Dorr \cite{bonnie2004bbn} implement a system using a combination of linguistically motivated sentence compression technique. Other notable methods for abstractive summarization include using discriminative tree-to-tree transduction model \cite{Cohn2008Sentence} and quasi-synchronous grammar approach utilizing both context-free parses and dependency parses \cite{woodsend2010generation}.

Recently, researchers have started utilizing deep learning framework in extractive and abstractive summarization. For extractive methods, Nallapati et al. \cite{nallapati2017summarunner} use recurrent neural networks (RNNs) to read the article and get the representations of the sentences and select important sentences. Yasunaga et al. \cite{yasunaga2017graph} combine RNNs with graph convolutional networks (CNNs) to compute the salience of each sentence. Narayan et al. \cite{narayan2018document} propose a framework composed of a hierarchical encoder based on CNNs and an attention-based extractor with attention over external information. More works are published recently on abstractive methods. Rush et al. \cite{rush2015neural} firstly apply modern neural networks to text summarization by using a local attention-based model to generate word conditioned on the input sentence. A bunch of work have been proposed to extend this approach, which achieving further improvements in performance. Chopra et al. \cite{chopra2016abstractive} use a similar convolutional attention-based encoder and replace the decoder with a conditional RNNs. Nallapati et al. \cite{nallapati2016abstractive} apply encoder-decoder RNNs framework with hierarchical attention and feature-rich embedding vector. Tan et al. \cite{tan2017abstractive} propose graph-based attention mechanism to summarize the salient information of document. However the above neural models cannot emit unseen words since the vocabulary is fixed at training time. In order to solve this problem, the point network \cite{vinyals2015pointer,see2017get} and the CopyNet \cite{gu2016incorporating} have been proposed to allow both copying words from the original text and generating words from a fixed vocabulary. Hsu et al. \cite{hsu2018unified} combine the strength of extractive and abstractive summarization and propose an inconsistency loss. Zhou et al. \cite{Zhou2017Selective} extend general encoder-decoder framework with a selective gate network, which helps improve encoding effectiveness and release the burden of the decoder. 

Our work has several significant improvements comparing with previous studies. First, to incorporate syntactic information, previous works only use unstructured linguistic information such as part-of-speech (POS) tags and named entity \cite{nallapati2016abstractive}. In this work, we utilize a structured syntactic parsing tree to learn a more effective context vector, which improves the performance of word prediction and alleviate the repetition problem. Second, to choose the salient information, previous works employ a selective gate network which is static during the decoding stage \cite{Zhou2017Selective}. We improve the gate network and let the states of the gate dynamically adjust according to the context of the decoder states, which is essential to document summarization.

% \cite{hsu2018unified} proposes a unified model combining the strength of extractive and abstractive summarization. Furthermore, \cite{paulus2017deep}\cite{narayan2018ranking} proposes some methods based on reinforcement learning (RL) to improve the quality of summary.

% Our work is inspired by the hybrid pointer-generator network of \cite{see2017get}, where it incorporates the copying operation into neural network-based Seq2Seq learning.

% , but there still have some differences. Below, we compare the related work with our solution on automatic text summarization: 1) Feature-rich encoder. Previous work only use unstructured linguistic information such as POS tags and named entity recognition, and simply concatenating these features into an embedding vector \cite{nallapati2016abstractive}. In this work, we utilize Syntactic Information such as constituency trees and dependency trees to learn a more effective context vector, which can improve the performance of word prediction and alleviate the repetition problem. 2) Dynamic Selective Gate. \cite{Zhou2017Selective} extends general encoder-decoder framework with a selective gate network, which helps improve encoding effectiveness and release the burden of the decoder. Our work upgrade their selective gate with Dynamic mechanism, which control the salient information flow from the static encoder to the current decoder to improve performance of word prediction. 

\begin{figure*}[!htbp]
	\centering
	\includegraphics[height=0.4\linewidth,width=0.8\linewidth]{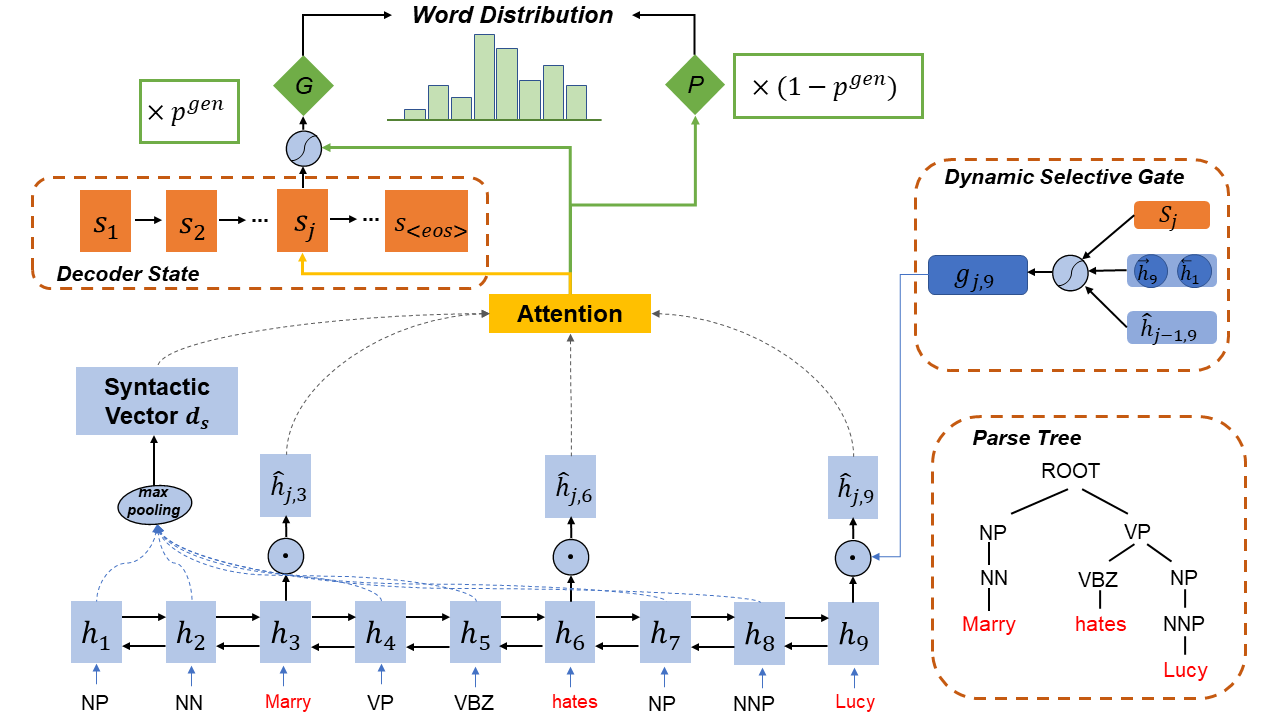}
	\caption{Overall architecture of the proposed syntactic and dynamic selective encoding model. The parsing tree of each sentence is serialized and fed into the encoder to help attain syntactic meanings. In $j$th decoding stage, the decoder benefits from the dynamic selective gate to drop out trivial words as well as attention mechanism influenced by the syntactic vector.}
	\label{fig:model}
\end{figure*}

\section{Methodology}
\label{sec:method}
In this section, we describe the proposed model. The architecture of the syntactic and dynamic selective syntactic encoding model is shown in Figure~\ref{fig:model}, which consists of the syntactic sequence encoder, the dynamic selective gates, and the pointer-generator network with syntactic attention decoder. 

\subsection{Syntactic Sequence Encoder}
%The goal of the sequence encoder is to transform the input document to a low-dimensional vector representation. 
%Previous studies treat a document as a sequence of words to build the source vector representation. These methods only utilize the semantic information but ignore the syntactic structure of the document.
%which treat a document as a sequence of words but ignore the syntactic structure of document.
Previous studies usually treat a document as a sequence of words but ignore the syntactic structure of document. To leverage the syntactic knowledge, we design a syntactic sequence encoder to learn document representations. 
%given a document, we first use a syntactic parser to build a parsing tree for each sentence in the document. The parsing tree of each sentence is serialized as a sequence and then concatenated into a unified neural network to build a document vector representation.
%we adopt Sequence Encoder with Source Syntax to encode the parse tree $l_{i}$ of sentence $s_{i}$ into the sentence representation, then concatenate sentences of a document $d$ into the document structure representation.
%The vector representation can capture not only the semantic information of the sentence, but also the syntactic structure of the corresponding parsing tree.

%Words in the parse tree $l_{i}$ are represented by leaf nodes while the non-leaf nodes are labeled by categories of the grammar. Non-leaf nodes represents the grammatical summarization of the sub-tree.
%failing to effectively employ the external knowledge. 
%\subsubsection{Word Encoder with Source Syntax}

A document $D$ is denoted as a sequence of sentences ${\bm q}$: $D=<{\bm q}_1, {\bm q}_2, ..., {\bm q}_n>$, where $n$ is the number of sentences in the document. For each sentence ${\bm q}_l$, we apply a syntactic parser to generate a parsing tree, and then adopt a depth-first traversal \cite{li2017modeling} to serialize the parsing tree to a sequence of tokens: ${\bm q}_l = <e_{1}, e_{2}, .., e_{k_l}>$, where $k_l$ is the number of tokens in the serialized parsing tree.
%$m$ is the length of a serialized parsing tree. 
Note that the token is not necessarily a word. In a parsing tree, a leaf node represents a word, while a non-leaf represents a parsing symbol including either a phrase label or a POS tag.

%For a document $d$ having a sequence of sentences $d = \{s_{1}, s_{2},.., s_{n}\}$, where $n$ is the sequence length of document, and a sentence $s_{i}$ having a sequence of the parse tree $l_{i}$, we adopts a depth-first traversal order of parse tree mixing both words and structural labels to get the sequence of the parse tree $l_{i} = \{l_{i,1},l_{i,2},..,l_{i,m}\}$, $m$ is the parser sequence length of sentence. Note that the leaf node $l_{i,k}$ is word, $l_{i,k}\in \upsilon_w$ and $\upsilon_w$ is the source vocabulary; the not-leaf node $l_{i,j}$ is structural label, $l_{i,j} \in \upsilon_l$ and $\upsilon_l$ is the structural labels vocabulary including phrase labels and POS tags.

%Given a document $d$ having linked parser sentences sequences $d = <l_{1},l_{2},..,l_{n}>$, 
Then, for a document, we concatenate all the serialized parsing trees into a long sequence $D' = <e_1, ..., e_m>$. Here $m$ is the total number of the tokens in all parsing trees $m=\sum_{l=1}^n k_l$. To model the sequential information, we first use an embedding vector ${\bm x}_i$ to represent the token $e_i$, which can be either a word or a symbol in the parsing tree. Then we employ a bidirectional long short-term memory (BiLSTM) \cite{graves2005framewise} to encoder the sequence information:
\begin{align}
&\overrightarrow{{\bm h}}_{i}=\overrightarrow{\text{LSTM}}({\bm x}_{i}), i\in\{1,..,m\}\\  
&\overleftarrow{{\bm h}}_{i}=\overleftarrow{\text{LSTM}}({\bm x}_{i}),i\in\{1,..,m\}
\end{align}
where $\overrightarrow{{\bm h}}_{i}$ and $\overleftarrow{{\bm h}}_{i}$ denote the hidden state of the forward LSTM and the backward LSTM, respectively. The whole representation of $i$th token is the concatenation of the hidden states from both directions ${\bm h}_i=[\overrightarrow{{\bm h}}_{i}, \overleftarrow{{\bm h}}_{m-i+1}]$.

To model the syntactic information, we apply the max-pooling over all the hidden states corresponding to the parsing symbols to produce the syntactic vector: ${\bm d}_s=\text{max\_pool}({\bm h}_{s}), s\in S$, where $S$ is the set of all parsing symbols in the document.

As shown in Figure~\ref{fig:model}, the syntactic sequence encoder takes the serialized parsing tree as the input. The BiLSTM  compute the hidden states for both the words (e.g. ``Mary"-${\bm h}_{3}$, ``hates"-${\bm h}_{6}$, ``Lucy"-${\bm h}_{9}$) and the parsing symbols (e.g. ``NP"-${\bm h}_{1}$, ``VP"-${\bm h}_{4}$, ``NNP"-${\bm h}_{8}$) as input. The word hidden states are used for further computation, while the hidden states of the parsing symbols are max-pooled to generate a syntactic vector ${\bm d}_s$.

%Syntactic vector will be fed help the pointer-generator network to select the salient information of the original text according to the structural information. We take max-pooling of annotation vectors of the structural labels to represent document syntactic vector $s_v$.

%\subsubsection{Sentence Encoder}
%Given a document $d$ having a sequence of sentences $d = \{s_{1},s_{2},..,s_{m}\}$, we also use a BiGRU to encode sentences:
%\begin{align}
%\overrightarrow{h_{i}}=\overrightarrow{GRU}(s_{i}), i\in\{1,..,n\}\\  
%\overleftarrow{h_{i}}=\overleftarrow{GRU}(s_{i}), i\in\{1,..,n\}
%\end{align}
%Where $s_{i}$ is the sentence $i$ vector introduced in section 3.2. The forward and backward hidden states are concatenated to get the document representation, $h_{i,j}=\{\overrightarrow{h_{i,j}},\overrightarrow{h_{i,j}}\}$.
  
\subsection{Dynamic Selective Gates}
As we discussed in Sec.\ref{sec:intro}, for document summarization, not all the information in the source should be fed into the decoder, and it is more important to only select the salient information and remove the unnecessary information from the input. Herein, we propose a novel dynamic selective gate to model the generation process of the salient information. We use a parameterized gate network to select the useful information for the summary generation. The gate state takes as input from both the state of the source and the previous state of the generated target, as well as a low-dimensional representation of the whole document ${\bm d}_h$, which is a concatenation of the last state of the forward LSTM and the first state of the backward LSTM ${\bm d}_h=[\overrightarrow{{\bm h}}_{m}, \overleftarrow{{\bm h}}_{1}]$.

Specifically, for the $i$th encoder step and $j$th decoder step, the state of the dynamic selective gate ${\bm g}_{j,i}$ is calculated as:
\begin{align}
\label{dGate_1}
&{\bm g}_{j, i}=\sigma({\bm w}^\top_g {\bm d}_h + {\bm u}^\top_g {\bm s}_{j} + {\bm v}^\top_g \hat{{\bm h}}_{j-1,i} + {\bm b}_g)  \\
&\hat{{\bm h}}_{j,i} = {\bm g}_{j,i} \odot {\bm h}_{i}
\end{align}
where ${\bm s}_j$ is the state of the $j$th step from the decoder LSTM which will be discussed in the next subsection, $\hat{\bm h}_{j-1,i}$ is the gated hidden state of the encoder, $\sigma$ denotes the sigmoid function, and $\odot$ denotes element-wise multiplication. When $j$ equals $0$, ~${\bm g}_{j,i}$ is set to 1s vector. ${\bm w}_g, {\bm u}_g, {\bm v}_g, {\bm b}_g$ are trainable parameters.

Note that, previous study \cite{Zhou2017Selective} has utilized the selective gate to control the information flow but the gate state only depends on the hidden states of the source text and the selective gate is static during the whole decoding stage. But the proposed dynamic selective gate depends on both the encoder and the decoder states, suggesting that the gate only open to the information which is useful for the \textit{current} target output rather than the whole target outputs. This is critical to document summarization, because the length of the document is long and a static gate may select much irrelevant information from the source at every decoding step. We will show the effectiveness of the dynamic selective gate in Sec.\ref{sec:case_analysis}.

\subsection{Pointer-generator Network with Syntactic Attention Decoder}
We use the recent proposed pointer-generator network \cite{see2017get} for decoding, which allows either copying words from the original text via pointers or generating new words based on the source vocabulary to handle the OOV problem.
%During the decoding stage, it is natural use the autoregressive model with the attention mechanism to generate the target. However, the abstractive text summarization may have the OOV problem during inference, where the salient information are not seen or rare in the source vocabulary during training. To handle the OOV problems, we adopt the recent proposed pointer-generator network \cite{see2017get}, which allows either copying words from the original text via pointers or generating new words based on the source vocabulary. 
%We combine the pointer-generator network with the syntactic and dynamic selective attention to generate the summarization.

Specifically, the attention strength between the $i$th source step and the $j$th target step is calculated by the current decoder state ${\bm s}_j$, the current gated encoder hidden state $\hat{{\bm h}}_{j,i}$, and the document syntactic vector ${\bm d}_s$. The context vector ${\bm c}_{j}$ is calculated by the attention-weighted summation of the gated encoding hidden states:
\begin{align}
&e_{j, i} = \tanh({\bm w}^\top_{a}\hat{{\bm h}}_{j,i}+{\bm u}_a^\top {\bm d}_s +{\bm v}_a^\top {\bm s}_{j} + {\bm b}_{a})\\
&a_{j, i}=\text{softmax}(e_{j, i})\\
&{\bm c}_{j}=\sum_{i}a_{j, i}\hat{{\bm h}}_{j,i}
\end{align}
where ${\bm w}_{a}$, ${\bm u}_{a}$, ${\bm v}_{a}$, ${\bm b}_{a}$ are trainable parameters.

An LSTM takes as input from the word embedding vector of the previous generated word ${\bm y}_{j-1}$, the previous context vector ${\bm c}_{j-1}$, and the previous decoder hidden state ${\bm s}_{j-1}$ to compute the new decoder state:
\begin{align}
&{\bm s}_{j} = \text{LSTM}({\bm y}_{j-1},{\bm c}_{j-1},{\bm s}_{j-1})
\end{align}
 and then the current context vector ${\bm c}_{j}$ and the current decoder hidden state ${\bm s}_{j}$ are fed into two linear layers and predicts the probability for each word $w$ in the vocabulary using the softmax function: 
\begin{align}
& p^{voc}_j(w)=\text{softmax}({\bm u}^\top_{v}({\bm w}^\top_{v}[{\bm c}_j,{\bm s}_j] + {\bm b}_{w}) + {\bm b}_{v})
\end{align}
where ${\bm w}_{v}$, ${\bm u}_{v}$, ${\bm b}_{w}$, ${\bm b}_{v}$ are trainable parameters.

% Where $W_a$, $U_a$, $V_a$ and $b_{a}$are learnable parameters. 

% At each decoding time step $t$, an LSTM receives the word embedding of the previous generated word $w_{t-1}$ and the previous context vector $c_{t-1}$ to compute the new decoder state $s_{t}$. The syntactic attention distribution $a_{t}=\{a_{t,1},..,a_{t,m}\}$ is calculated by the current decoder state $s_{t}$, the current selected encoder hidden state $H_{t}^{*}$ and the document structural vectors $sv$. The syntactic attention represents the importance score of the current selected encoder hidden states $H_{t}^{*}$ and is normalized to get the current context vector $c_{t}$ by weighted sum:
% \begin{align}
% &s_{t} = LSTM(w_{t-1},c_{t-1},s_{t-1})	\\
% &e_{t,j} = v^\top tanh(W_{a}h^{*}_{t,j}+U_{a}sv +V_{a}s_{t}+b_{a})\\
% &a_{t}=softmax(e_{t})\\
% &c_{t}=\sum_{i=1}^{n}\sum_{j=1}^{m}a_{t,j}h^{*}_{t,j}
% \end{align}
% Where $W_a$, $U_a$, $V_a$ and $b_{a}$are learnable parameters. 

% The context vector $c_{t}$ and the current decoder state $s_t$ are concatenated to pass two linear layers and predict the next word with a $softmax$ layer:
% \begin{align}
% P_{vocab} = softmax(V_v(W_v[c_{t},s_{t}]+b_w)+b_{v})	
% \end{align}

Further, a pointer-generator network produces the switch probability $p^{gen}$ to decide whether generates a word by $p^{voc}$ or copies a word from the original source text. $p^{gen}$ is calculated from the context vector ${\bm c}_{j}$, the decoder state ${\bm s}_{j}$ and the decoder word ${\bm y}_{j}$. The final probability with the word $w$ is calculated based on $p^{gen}$ and the attention distribution:
\begin{align}
&p^{gen} = \sigma({\bm w}_{p}^{\top}{\bm c}_{j}+{\bm u}_{p}^{\top}{\bm s}_{j}+{\bm v}_{p}^{\top}{\bm y}_{j}+{\bm b}_{p})	\\
&p(w) = p^{gen}p^{voc}(w) + (1-p^{gen})\sum_{i\in \{e_i=w\}}{a}_{j, i}
\end{align}
where ${\bm w}_{p}$, ${\bm u}_{p}$, ${\bm v}_{p}$, ${\bm b}_{p}$ are trainable parameters.

%Next, pgen is used as a soft switch to choose between generating a word from the vocabulary by sampling from $Pvocab$, or copying a word from the input sequence by 

%sampling from the attention distribution. For each document let the extended vocabulary denote the union of the vocabulary, and all words appearing in the source document. We obtain the following probability distribution over the extended vocabulary:
\subsection{Model Training}
To train the model, we use the negative log-likelihood function $-\log p({\bm y}|D)$ as the loss for each document. We further adopt the coverage loss from See et al. \cite{see2017get}, aiming to handle the repetition problem in text summarization. The coverage loss at the decoding step $j$ corresponding to the encoding step $i$ is the summation of attention
distributions over all previous decoding step: ${\hat a}_{j,i}=\sum_{t=0}^{j-1}{a}_{t,i}$. The final loss function at the decoding step $j$ is:
\begin{align}
\label{loss_2}
\mathcal{L}({\bm y}_j) = - \log p({\bm y}_j|D) + \lambda\sum_{i}\min(a_{j,i},{\hat a}_{j,i})
 \end{align}
 
% \begin{align}
% \label{loss_1}
% loss = - \sum_{(d,y)\in D}\log p({\bm y}|D)
% \end{align}

% The loss function (\ref{loss_1}) of the model is to maximize the output summary probability given the input document. Therefore, we optimize the negative log-likelihood loss function:
% \begin{align}
% \label{loss_1}
% loss = - \frac{1}{|D|} \sum_{(d,y)\in D}\log p(y|d)
% \end{align}

% Furthermore, to handle repetition problem, we adopt the coverage mechanism \cite{see2017get}, which adds the coverage vector $cv_{t}=\sum_{t^{'}=0}^{t-1}a_{t^{'}}$ to the attention mechanism and the coverage loss to loss function(\ref{loss_1}) to penalize for selecting the same encoder information repeatedly:
% \begin{align}
% \label{loss_2}
% loss = - \frac{1}{|D|} \sum_{(d,y)\in D}(\log p(y|d) + \lambda\sum_{t}min(a_{t,i},cv_{t,i}))
%  \end{align}

\section{Experiments}
\label{sec:exp}
In this section, we describe the experiment details including datasets, implementation details, baselines and the results. 

\subsection{Datasets}
We conduct experiments on CNN/Daily Mail\footnote{https://github.com/abisee/pointer-generator} dataset \cite{see2017get}, which comprises multi-sentence summaries and has been widely used in automatic text summarization. We use released scripts\footnote{https://github.com/abisee/cnn-dailymail} to obtain the same version of the the data, which has 287,227 training pairs, 13,368 validation pairs and 11,490 test pairs. The source documents have 681 words spanning 40 sentences on an average while the summaries consist of 48 words and 3.9 sentences. The dataset is released in two versions: one is anonymized version which has been pre-processed to replace each named entity, and the other is the original version consisting of actual entity names. In this work, we use the original text since it requires no pre-processing and is more challenging because anonymized dataset replaces named entities with unique identifier, which always are out of vocabulary. In the following experiments all the models are trained and tested with three different datasets separately, including CNN corpus, Daily Mail corpus and the combination of CNN and Daily Mail corpus. Table~\ref{tab:dataset} shows the detail statistics information of experiment datasets. 
\begin{table}[htbp]
\caption{Data statistics for CNN and CNN/Daily Mail datasets. AvgDocSents is the average sentences number of original documents and AvgDocWords is the average sentences length of original documents. AvgSumSents is the average sentences number of summaries and AvgDocWords is the average sentences length of summaries.}
\scalebox{0.8}{
 \begin{tabular}{cccc}
  \toprule
  Data Set&CNN &Daily &CNN/Daily \\
  \midrule
 AvgDocSents& 34.4&42&40\\
 AvgDocWords&655&692&681\\
 AvgSumWords &3.7&4&3.92 \\
 AvgSumSents&42&42&48.3\\
  \bottomrule
 \end{tabular}}
 \label{tab:dataset}
\end{table}

\subsection{Implementation}
For all experiments, we use 50k words of the source vocabulary and Stanford Constituency Parser\footnote{https://nlp.stanford.edu/software/srparser.html} to get the syntactic information of the sentences in the corpora, which includes 16 phrase labels and 32 POS tags. Our model takes 256-dimensional hidden states, 128-dimensional word embedding vectors and use adagrad with learning rate 0.15 and initialize the accumulator value with 0.1. This was found to work best among stochastic gradient descent, adadelta, momentum, adam and RMSprop. In addition, we set the maximum length of sentence on source-side to 1200, on target-side for training and testing to 100 and 120 respectively. To both decode fast and get better results, we set the beam size to 4 in our experiments. Furthermore, we added the coverage mechanism in loss function (\ref{loss_2}) with coverage loss weighted to $\lambda = 1$.

\subsection{Baselines}
We compare our proposed model with several state-of-the-art automatic text summarization systems and techniques consisting of extractive and abstractive methods.\footnote{The results of baselines are incomplete on sub-dataset because some other researchers chose to report results on only one sub-dataset;}:
\begin{itemize}[itemsep=2pt,parsep=2pt]
\item \textbf{Lead-3} is a standard extractive baseline, which generates summary simply by selecting the "leading" three sentences from source document. 
\item \textbf{NN-SE} \cite{Cheng2016Neural} utilizes encoder-decoder framework, which learns the representation of source though encoder and classifies sentences of document by decoder.
\item \textbf{SummaRuNN} \cite{nallapati2016abstractive} applies encoder-decoder RNN abstractive framework with hierarchical attention and feature-rich embedding vector.
\item \textbf{SummaRuNNer}\cite{nallapati2017summarunner} treats extractive summarization as a sequence classification problem, where a binary decision has been made on each sentence about whether or not it should be included in the summary. 
\item \textbf{SummaRuNNer-abs}\cite{nallapati2017summarunner} is also an extractive model similar to SummaRuNNer but is trained directly on the abstractive summaries.
\item \textbf{Seq2Seq+attn} \cite{see2017get} We use a Seq2Seq framework based on Uni-GRU with non-hierarchical attention as our baseline model. 
\item \textbf{Distraction-M3} \cite{chen2016distraction} is an extension of Seq2Seq+attn model with distract mechanism to traverse between different content of a document to better grasp the overall meaning for summarization.
\item \textbf{Graph-Based Model} \cite{tan2017abstractive} proposes a novel abstractive graph-based attention mechanism in the Seq2Seq framework, which aims to find salient content from the original document.
\item \textbf{DeepRL} \cite{paulus2017deep} proposes a unified framework combining Seq2Seq and RL into to improve the quality of summary.
\item \textbf{Pointer-generator+Coverage}(Po-Gen+Cov)\cite{see2017get} improves the standard Seq2Seq model with a hybrid pointer-generator, which can not only produce novel words but also copy words from the source text.
\item \textbf{SelectiveGate} \cite{Zhou2017Selective} proposes the encoder-decoder framework based on a static selective gate network, which helps improve encoding effectiveness and release the burden of the decoder.
\end{itemize}

\subsection{Experimental Results}
We adopt the widely used ROUGE\cite{lin2004rouge} by pyrouge \footnote{pypi.python.org/pypi/pyrouge/0.1.3} for evaluation metric. It measures the similarity of the output summary and the standard reference by computing overlapping n-gram, such as unigram, bigram and longest common subsequence. In the following experiments, we adopt ROUGE-1 (unigram), ROUGE-2 (bigram) and ROUGE-L (longest common subsequence) for evaluation.

It can be observed from Tables \ref{tab:cnn} and \ref{tab:cnn/dailymail} that the proposed approach achieves the best performance on the two datasets. Our best model outperforms all baseline extractive and abstractive models on ROUGE-1, ROUGE-2 and ROUGE-L. Compared with abstractive Graph-based, RL-based and SummaRunner model, our model leverages the structural information of document and improves the pointer-network with syntactic attention to copy relevant words in semantic and structural aspect from the original text to handle OOV problems, while Graph-based, RL-based and SummaRunner model all take the anonymized data, which has replaced named entity with ``@entity" to alleviate OOV problems. Furthermore, unlike Graph-based, RL-based and SummaRunner model, we do not pretrain the word embedding vectors.

\begin{table}[htbp]
\caption{Comparison results on the CNN test set respectively using the full-length F1 variants of Rouge. Baseline model results with $^{\ddagger}$ mark are taken from the corresponding papers. All our ROUGE scores have a 95\% confidence interval of at most $\pm$0.25 as reported by the official ROUGE
script.}
\scalebox{0.8}{
 \begin{tabular}{cccc}
  \toprule
  Method & R1 &R2 &RL \\ 
  \midrule
%    Lead-3$^{\ddagger}$&26.1&9.6&17.8\\
    Seq2Seq+attn$^{\ddagger}$ & 18.4 & 4.8 & 14.3\\
    Distraction-M3$^{\ddagger}$ & 27.1 & 8.2 & 18.7\\
    Graph-based model$^{\ddagger}$ & 30.3&9.8 &20.0\\
  	Po-Gen+Cov &29.8&10.4&26.6\\
  	SelectiveGate(w/o Po-Gen+Cov) &19.8&5.8&15.6\\
  	SelectiveGate(Po-Gen+Cov) &30.1&10.6&26.5\\
  \midrule
   	Our Model(Source Syntax)&30.6 &11.1&26.8\\
  	Our Model(Dynamic Selective)&30.7&10.9&26.9\\ 	
   Our Model(All) &\textbf{31.2}&\textbf{11.5}&\textbf{27.3}\\
  \bottomrule
 \end{tabular}}
 \label{tab:cnn}
\end{table}

\begin{table}[htbp]
\begin{center}
	\caption{Comparison results on the CNN/Daily Mail test set using the full-length F1 variants of Rouge. Baseline model results with $^{\ddagger}$ mark are taken from the corresponding papers. 150k represents vocabulary size of 150k and 50k represents vocabulary size of 50k. All our ROUGE scores have a 95\% confidence interval of at most $\pm$0.25 as reported by the official ROUGE.}
\scalebox{0.8}{
 \begin{tabular}{cccc}
  \toprule
  	Method & R1 &R2 &RL\\ 
  \midrule
    Seq2Seq+attn(150k)$^{\ddagger}$ & 30.49 & 11.17 & 28.08\\
    Seq2Seq+attn(50k)$^{\ddagger}$ & 31.33 & 11.81 & 28.83\\
    SummaRuNN$^{\ddagger}$  & 35.46 & 13.30 & 32.65\\
    SummaRuNNer$^{\ddagger}$ & 39.6&16.2 &35.3\\
    Graph-based model $^{\ddagger}$ & 38.1&13.9 &34.0\\
    DeepRL $^{\ddagger}$ & 39.87&15.82 &36.90\\
    Pointer-generator+coverage & 39.53&17.28 &36.38\\
  \midrule
  Our model & \textbf{40.37} & \textbf{17.82} & \textbf{37.3} \\
  \midrule
  lead-3$^{\ddagger}$ &40.34 &17.70&36.57\\
  \bottomrule
 \end{tabular}}
 \label{tab:cnn/dailymail}
\end{center}
\end{table}

% It can be observed from Tables \ref{tab:cnn} and \ref{tab:cnn/dailymail} that the proposed approach achieves the best performance on the two datasets. Our best model outperforms all baseline extractive and abstractive models on ROUGE-1, ROUGE-2 and ROUGE-L. Compared with abstractive Graph-based, RL-based and SummaRunner model, our model leverages the structural information of document and improves the pointer-network with syntactic attention to copy relevant words in semantic and structural aspect from the original text to handle OOV problems, while Graph-based, RL-based and SummaRunner model all take the anonymized data, which has replaced named entity with "@entity" to alleviate OOV problems. Furthermore, unlike Graph-based, RL-based and SummaRunner model, we do not pretrain the word embedding vectors.

We also compare in detail with two similar methods in Table \ref{tab:cnn}. For Pointer-generator with coverage (Po-Gen+Cov) model, we show that, with the help of structural information and dynamic selective gate, the scores of our best model performs the best over the Po-Gen+Cov model on evaluation metrics (1.4 ROUGE-1, 1.1 ROUGE-2 and 0.7 ROUGE-L). For static SelectiveGate model, we conduct two experiments with Po-Gen+Cov and without Po-Gen+Cov due to its original paper focusing on short-text summarization, which does not use Po-Gen+Cov mechanism to alleviate OOV and word repetition problems. The result demonstrates that the static SelectiveGate improves the performance of Po-Gen+Cov model and the dynamic SelectiveGate can further improve the ROUGE scores of static SelectiveGate model by selecting current important information for decoding in every time step.
 
Further, to study the different impacts of source syntax and dynamic selective gate on the performance of the proposed model, we conduct ablation experiments on the CNN dataset, where we train the model with the source syntax encoding only and the dynamic selective gate only, respectively. As shown in the last three rows in Table \ref{tab:cnn}, that 1) either the source syntax encoding or the dynamic selective gate can improve the performance compared with SelectiveGate approach; 2) combining both approach leads to further improvement which achieves the best results as shown in the last row in the table.

\begin{table}[htbp]
\begin{center}
\caption{Comparison results with different document lengths on the CNN dataset respectively using the full-length F1 variants of Rouge. All our ROUGE scores have a 95\% confidence interval of at most $\pm$0.25 as reported by the official ROUGE.}
\scalebox{0.8}{
 \begin{tabular}{cccc}
  \toprule
  L & R1 &R2 &RL \\
  \midrule
    1000&31.1&11.4&27.1\\
    1200&31.2&11.5&27.3\\
    1400&31.0&11.3&27.2\\
\bottomrule
 \end{tabular}}
 \label{tab:length}
 \end{center}
\end{table}

In addition, to study the impact of the lengths of document on the performance of the proposed model, we conduct experiments on the CNN test sets between 1000 and 1400. Table \ref{tab:length} clearly shows that the performance of the proposed approach is stable across different lengths of document.

\subsection{Example Analysis}
\label{sec:case_analysis}
In this subsection, we use an example to show the effectiveness of the syntactic encoding and the dynamic selective mechanism. We choose the same example introduced in Figure \ref{fig:example}. 

\begin{figure}[h]
	%\vspace{-2ex}
	\centering
	\includegraphics[width=\linewidth]{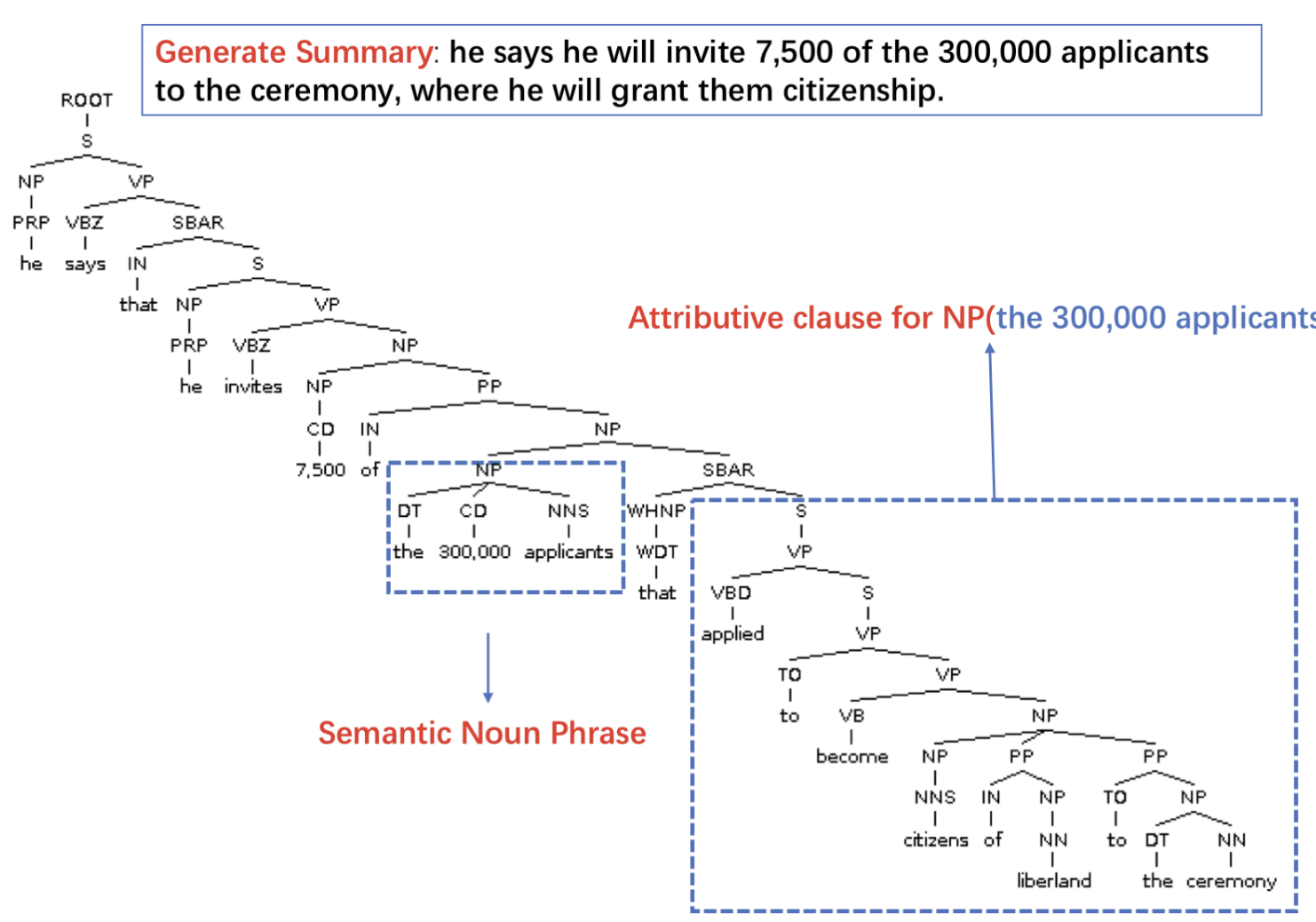}
	%\vspace{-2ex}
	\caption{A parsing tree corresponding to the blue text in Figure \ref{fig:example}. The dashed box shows that `` applied to .. ceremony" is an attributive clause for ``300, 000 applicants". The upper box shows the our generated summary which correctly summarize the original document.}
	%\vspace{-2ex}
	\label{fig:syntactic}
\end{figure}

Figure \ref{fig:syntactic} shows the constituency parsing tree of the sentence. The baseline model in Figure \ref{fig:example} generates wrong summary because it is not able to model the syntactic structure, e.g., ``300, 000 applicants" is a noun phase, and ``applied to .. ceremony" is an attributive clause. In our approach, the BiLSTM encoder takes as input from the serialized parsing tree, and each word token is surrounded by the parsing symbols. Intuitively, if two consecutive words do not belong to the same syntactic subtree, more parsing symbols will be inserted between them in the BiLSTM encoder and it will be less likely that these two symbols have strong connection. As shown at the top in Figure \ref{fig:syntactic}, the generated summary of our model correctly conveys the summary of the source text. 

\begin{figure}[h]
	%\vspace{-2ex}
	\centering
	\includegraphics[width=\linewidth]{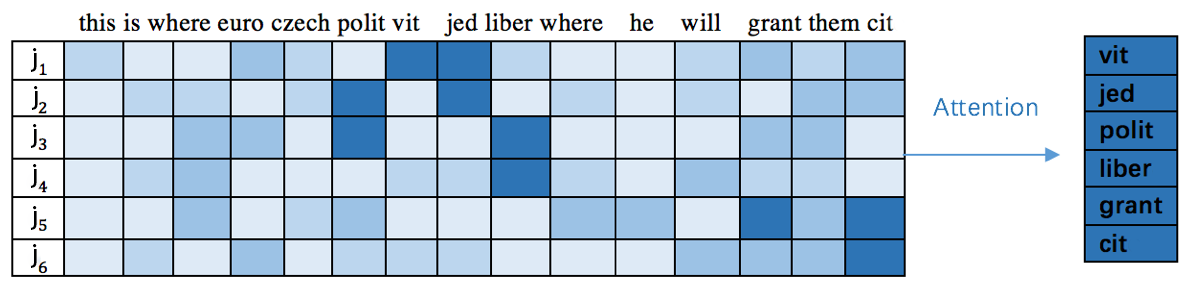}
	%\vspace{-2ex}
	\caption{Visualization of the dynamic selective gates. The gates dynamically adjust the states in different decoding steps $j_1,...j_6$. Darkness of the blocks indicates the openness of the gates.}
	%\vspace{-2ex}
	\label{fig:selective}
\end{figure}

For the dynamic selective gate, we use a method in \cite{li2015visualizing} to visualize it. The method defines a highly non-linear function to measure the contribution of the source word $w_i$ gated by ${\bm g}_{j,i}$ in the $j$th generation step. As shown in Figure \ref{fig:selective}, the dynamic selective mechanism can select the most important information from the original document in every decoding step. For example, at decoding step $j_{1}$, the selective gate filters out some nonsensical words (e.g. ``the", ``is", ``he") and selects current important words (e.g. ``jedlicka", ``politician") to help the following attention to generate the most important word (e.g. ``jedlicka"). Furthermore, Figure \ref{fig:selective} also shows that the word out of source vocabulary can also be generated (e.g. ``vit", ``jedlicka") but the weight of the selected words will decrease in the next decoding steps, indicating that our model can address the OOV problem and the word repetition problem.

\section{Conclusions and Future Work}
\label{sec:con_fut}
In this work, we propose a novel document summarization model, which takes its input from a serialized parsing tree which enables the encoder to learn the inherent syntactic structure from each sentence in the document. Further, we propose a dynamic selective gate to control the information flow from the encoder to the decoder. This mechanism  dynamically control the salient information based on the context of the decoder state, which is essential to document summarization. The experimental results on two long-text datasets CNN/Daily Mail show the advantage of our model over several baseline approaches. 

In this work, we use the depth-first traversal to generate the serialized parsing tree, which is not able to contain all structure information from the tree. We will consider better representation to model the hierarchical structure of the parsing tree. On the other hand, the proposed dynamic selective gate only apply to the word tokens in the source. We may consider integrating both the word tokens and the parsing symbol to produce better the information flow from the source to the target.

\bibliographystyle{IEEEtran}
\bibliography{selective_tree_summary_ijcnn}

\begin{thebibliography}{10}
\providecommand{\url}[1]{#1}
\csname url@samestyle\endcsname
\providecommand{\newblock}{\relax}
\providecommand{\bibinfo}[2]{#2}
\providecommand{\BIBentrySTDinterwordspacing}{\spaceskip=0pt\relax}
\providecommand{\BIBentryALTinterwordstretchfactor}{4}
\providecommand{\BIBentryALTinterwordspacing}{\spaceskip=\fontdimen2\font plus
\BIBentryALTinterwordstretchfactor\fontdimen3\font minus
  \fontdimen4\font\relax}
\providecommand{\BIBforeignlanguage}[2]{{%
\expandafter\ifx\csname l@#1\endcsname\relax
\typeout{** WARNING: IEEEtran.bst: No hyphenation pattern has been}%
\typeout{** loaded for the language `#1'. Using the pattern for}%
\typeout{** the default language instead.}%
\else
\language=\csname l@#1\endcsname
\fi
#2}}
\providecommand{\BIBdecl}{\relax}
\BIBdecl

\bibitem{sutskever2014sequence}
I.~Sutskever, O.~Vinyals, and Q.~V. Le, ``Sequence to sequence learning with
  neural networks,'' in \emph{Proceedings of the 28th Conference on Neural
  Information Processing Systems}, 2014, pp. 3104--3112.

\bibitem{bahdanau2014neural}
D.~Bahdanau, K.~Cho, and Y.~Bengio, ``Neural machine translation by jointly
  learning to align and translate,'' \emph{arXiv preprint arXiv:1409.0473},
  2014.

\bibitem{luong2015effective}
T.~Luong, H.~Pham, and C.~D. Manning, ``Effective approaches to attention-based
  neural machine translation,'' in \emph{Proceedings of the 2015 Conference on
  Empirical Methods in Natural Language Processing}, 2015, pp. 1412--1421.

\bibitem{serban2016building}
I.~V. Serban, A.~Sordoni, Y.~Bengio, A.~Courville, and J.~Pineau, ``Building
  end-to-end dialogue systems using generative hierarchical neural network
  models,'' in \emph{Proceedings of the 30th AAAI Conference on Artificial
  Intelligence}.\hskip 1em plus 0.5em minus 0.4em\relax AAAI Press, 2016, pp.
  3776--3783.

\bibitem{bordes2016learning}
A.~Bordes, Y.-L. Boureau, and J.~Weston, ``Learning end-to-end goal-oriented
  dialog,'' \emph{arXiv preprint arXiv:1605.07683}, 2016.

\bibitem{gu2016incorporating}
J.~Gu, Z.~Lu, H.~Li, and V.~O. Li, ``Incorporating copying mechanism in
  sequence-to-sequence learning,'' \emph{arXiv preprint arXiv:1603.06393},
  2016.

\bibitem{nallapati2016abstractive}
R.~Nallapati, B.~Zhou, C.~dos Santos, C.~Gulcehre, and B.~Xiang, ``Abstractive
  text summarization using sequence-to-sequence rnns and beyond,'' in
  \emph{Proceedings of The 20th SIGNLL Conference on Computational Natural
  Language Learning}, 2016, pp. 280--290.

\bibitem{see2017get}
A.~See, P.~J. Liu, and C.~D. Manning, ``Get to the point: Summarization with
  pointer-generator networks,'' in \emph{Proceedings of the 55th Annual Meeting
  of the Association for Computational Linguistics (Volume 1: Long Papers)},
  vol.~1, 2017, pp. 1073--1083.

\bibitem{tan2017abstractive}
J.~Tan, X.~Wan, and J.~Xiao, ``Abstractive document summarization with a
  graph-based attentional neural model,'' in \emph{Proceedings of the 55th
  Annual Meeting of the Association for Computational Linguistics (Volume 1:
  Long Papers)}, vol.~1, 2017, pp. 1171--1181.

\bibitem{li2017modeling}
J.~Li, D.~Xiong, Z.~Tu, M.~Zhu, M.~Zhang, and G.~Zhou, ``Modeling source syntax
  for neural machine translation,'' in \emph{Proceedings of the 55th Annual
  Meeting of the Association for Computational Linguistics (Volume 1: Long
  Papers)}, vol.~1, 2017, pp. 688--697.

\bibitem{Zhou2017Selective}
Q.~Zhou, N.~Yang, F.~Wei, and M.~Zhou, ``Selective encoding for abstractive
  sentence summarization,'' in \emph{Proceedings of the 55th Annual Meeting of
  the Association for Computational Linguistics (Volume 1: Long Papers)},
  vol.~1, 2017, pp. 1095--1104.

\bibitem{Wong2008Extractive}
K.-F. Wong, M.~Wu, and W.~Li, ``Extractive summarization using supervised and
  semi-supervised learning,'' in \emph{Proceedings of the 22nd International
  Conference on Computational Linguistics-Volume 1}.\hskip 1em plus 0.5em minus
  0.4em\relax Association for Computational Linguistics, 2008, pp. 985--992.

\bibitem{Wang2009Multi}
D.~Wang, S.~Zhu, T.~Li, and Y.~Gong, ``Multi-document summarization using
  sentence-based topic models,'' in \emph{Proceedings of the 47th Annual
  Meeting of the Association for Computational Linguistics}.\hskip 1em plus
  0.5em minus 0.4em\relax Association for Computational Linguistics, 2009, pp.
  297--300.

\bibitem{celikyilmaz2010hybrid}
A.~Celikyilmaz and D.~Hakkani-Tur, ``A hybrid hierarchical model for
  multi-document summarization,'' in \emph{Proceedings of the 48th Annual
  Meeting of the Association for Computational Linguistics}.\hskip 1em plus
  0.5em minus 0.4em\relax Association for Computational Linguistics, 2010, pp.
  815--824.

\bibitem{Alguliev2011MCMR}
R.~M. Alguliev, R.~M. Aliguliyev, M.~S. Hajirahimova, and C.~A. Mehdiyev,
  ``Mcmr: Maximum coverage and minimum redundant text summarization model,''
  \emph{Expert Systems with Applications}, vol.~38, no.~12, pp.
  14\,514--14\,522, 2011.

\bibitem{Erkan2011LexRank}
G.~Erkan and D.~R. Radev, ``Lexrank: Graph-based lexical centrality as salience
  in text summarization,'' \emph{Journal of artificial intelligence research},
  vol.~22, pp. 457--479, 2004.

\bibitem{Alguliev2013Multiple}
R.~M. Alguliev, R.~M. Aliguliyev, and N.~R. Isazade, ``Multiple documents
  summarization based on evolutionary optimization algorithm,'' \emph{Expert
  Systems with Applications}, vol.~40, no.~5, pp. 1675--1689, 2013.

\bibitem{banko2000headline}
M.~Banko, V.~O. Mittal, and M.~J. Witbrock, ``Headline generation based on
  statistical translation,'' in \emph{Proceedings of the 38th Annual Meeting on
  Association for Computational Linguistics}.\hskip 1em plus 0.5em minus
  0.4em\relax Association for Computational Linguistics, 2000, pp. 318--325.

\bibitem{bonnie2004bbn}
D.~Z. Bonnie and B.~Dorr, ``Bbn/umd at duc-2004: Topiary,'' in
  \emph{Proceedings of the 2004 Document Understanding Conference (DUC 2004) at
  Conference of the North American Chapter of the Association for Computational
  Linguistics: Human Language Technologies}.\hskip 1em plus 0.5em minus
  0.4em\relax Citeseer, 2004.

\bibitem{Cohn2008Sentence}
T.~Cohn and M.~Lapata, ``Sentence compression beyond word deletion,'' in
  \emph{Proceedings of the 22nd International Conference on Computational
  Linguistics-Volume 1}.\hskip 1em plus 0.5em minus 0.4em\relax Association for
  Computational Linguistics, 2008, pp. 137--144.

\bibitem{woodsend2010generation}
K.~Woodsend, Y.~Feng, and M.~Lapata, ``Generation with quasi-synchronous
  grammar,'' in \emph{Proceedings of the 2010 conference on empirical methods
  in natural language processing}.\hskip 1em plus 0.5em minus 0.4em\relax
  Association for Computational Linguistics, 2010, pp. 513--523.

\bibitem{nallapati2017summarunner}
R.~Nallapati, F.~Zhai, and B.~Zhou, ``Summarunner: A recurrent neural network
  based sequence model for extractive summarization of documents.'' in
  \emph{Proceedings of the 31st AAAI Conference on Artificial Intelligence},
  2017, pp. 3075--3081.

\bibitem{yasunaga2017graph}
M.~Yasunaga, R.~Zhang, K.~Meelu, A.~Pareek, K.~Srinivasan, and D.~Radev,
  ``Graph-based neural multi-document summarization,'' \emph{arXiv preprint
  arXiv:1706.06681}, 2017.

\bibitem{narayan2018document}
S.~Narayan, R.~Cardenas, N.~Papasarantopoulos, S.~B. Cohen, M.~Lapata, J.~Yu,
  and Y.~Chang, ``Document modeling with external attention for sentence
  extraction,'' in \emph{Proceedings of the 56th Annual Meeting of the
  Association for Computational Linguistics (Volume 1: Long Papers)}, vol.~1,
  2018, pp. 2020--2030.

\bibitem{rush2015neural}
A.~M. Rush, S.~Chopra, and J.~Weston, ``A neural attention model for
  abstractive sentence summarization,'' in \emph{Proceedings of the 2015
  Conference on Empirical Methods in Natural Language Processing}, 2015, pp.
  379--389.

\bibitem{chopra2016abstractive}
S.~Chopra, M.~Auli, and A.~M. Rush, ``Abstractive sentence summarization with
  attentive recurrent neural networks,'' in \emph{Proceedings of the 15th
  Annual Conference of the North American Chapter of the Association for
  Computational Linguistics: Human Language Technologies}, 2016, pp. 93--98.

\bibitem{vinyals2015pointer}
O.~Vinyals, M.~Fortunato, and N.~Jaitly, ``Pointer networks,'' in
  \emph{Proceedings of the Twenty-ninth Conference on Neural Information
  Processing Systems}, 2015, pp. 2692--2700.

\bibitem{hsu2018unified}
W.-T. Hsu, C.-K. Lin, M.-Y. Lee, K.~Min, J.~Tang, and M.~Sun, ``A unified model
  for extractive and abstractive summarization using inconsistency loss,'' in
  \emph{Proceedings of the 56th Annual Meeting of the Association for
  Computational Linguistics (Volume 1: Long Papers)}, 2018.

\bibitem{graves2005framewise}
A.~Graves and J.~Schmidhuber, ``Framewise phoneme classification with
  bidirectional lstm and other neural network architectures,'' \emph{Neural
  Networks}, vol.~18, no. 5-6, pp. 602--610, 2005.

\bibitem{Cheng2016Neural}
J.~Cheng and M.~Lapata, ``Neural summarization by extracting sentences and
  words,'' in \emph{Proceedings of the 54th Annual Meeting of the Association
  for Computational Linguistics (Volume 1: Long Papers)}, vol.~1, 2016, pp.
  484--494.

\bibitem{chen2016distraction}
Q.~Chen, X.~Zhu, Z.~Ling, S.~Wei, and H.~Jiang, ``Distraction-based neural
  networks for modeling documents,'' in \emph{Proceedings of the 25th
  International Joint Conference on Artificial Intelligence}.\hskip 1em plus
  0.5em minus 0.4em\relax AAAI Press, 2016, pp. 2754--2760.

\bibitem{paulus2017deep}
R.~Paulus, C.~Xiong, and R.~Socher, ``A deep reinforced model for abstractive
  summarization,'' \emph{arXiv preprint arXiv:1705.04304}, 2017.

\bibitem{lin2004rouge}
C.-Y. Lin, ``Rouge: A package for automatic evaluation of summaries,''
  \emph{Text Summarization Branches Out}, 2004.

\bibitem{li2015visualizing}
J.~Li, X.~Chen, E.~Hovy, and D.~Jurafsky, ``Visualizing and understanding
  neural models in nlp,'' \emph{arXiv preprint arXiv:1506.01066}, 2015.

\end{thebibliography}

\end{document}